\documentclass{article} 
\usepackage{iclr2027_conference}
\usepackage{times}

\usepackage[T1]{fontenc}
\usepackage[english]{babel}
\usepackage{microtype}
\usepackage{amsmath}
\usepackage{amssymb}
\usepackage{amsfonts}
\usepackage{bm}
\usepackage{hyperref}
\usepackage[nameinlink]{cleveref}
\usepackage{url}
\usepackage[dvipsnames]{xcolor}
\usepackage{graphicx}
\usepackage[printonlyused]{acronym}
\usepackage{enumitem}
\usepackage{booktabs}
\usepackage{algorithm}
\usepackage{algpseudocode}
\usepackage{wrapfig}

\crefname{table}{Tab.}{Tabs.}
\Crefname{table}{Tab.}{Tabs.}

\crefname{figure}{Fig.}{Figs.}
\Crefname{figure}{Fig.}{Figs.}

\crefname{equation}{Eq.}{Eqs.}
\Crefname{equation}{Eq.}{Eqs.}

\crefname{algorithm}{Alg.}{Algs.}
\Crefname{algorithm}{Alg.}{Algs.}

\newcommand{\methodname}{PT-Denoise}

\crefname{appendix}{appendix}{appendices}
\AddToHook{cmd/appendix/before}{\crefalias{section}{appendix}}

\newacro{CO}{combinatorial optimization}
\newacro{GNN}[GNN]{graph neural network}
\newacro{MIS}{maximum independent set}
\newacro{MDS}{minimum dominating set}
\newacro{MCMC}{Markov chain Monte Carlo}
\newacro{BA}{Barabási--Albert}
\newacro{LLM}{large language model}

\newcommand{\meanstd}[2]{#1{\footnotesize$\pm$#2}}

\DeclareMathOperator*{\argmin}{arg\,min}
\DeclareMathOperator{\Cat}{Cat}
\DeclareMathOperator{\softmax}{softmax}

\title{Parallel Tempering for Diffusion-Based \\ Combinatorial Optimization}

\author{%
    Arman Mielke$^{1,2,3}$\thanks{Correspondence to \texttt{arman.mielke@etas.com}} \quad
    Uwe Bauknecht$^1$ \quad
    Thilo Strauss$^4$ \quad
    Mathias Niepert$^{2,3,5}$ \\
    $^1$ETAS Research \qquad
    $^2$University of Stuttgart \\
    $^3$Max Planck Research School for Intelligent Systems (IMPRS-IS) \\
    $^4$School of AI and Advanced Computing, Xi'an Jiaotong-Liverpool University \\
    $^5$NEC Laboratories Europe
}

\iclrfinalcopy 

\begin{document}

\maketitle
\lhead{Preprint. Under review.}

\begin{abstract}
    Discrete diffusion models have emerged as a powerful paradigm for solving combinatorial optimization (CO) problems on graphs by learning to sample high-quality solutions.
    A common inference-time approach is to generate multiple candidate solutions independently and return the best-performing sample, improving solution quality at the expense of an increase in computational cost.
    In this work, we introduce PT-Denoise, an inference-time procedure that allows these concurrent denoising trajectories to interact through parallel tempering, without requiring retraining or fine-tuning of the underlying denoiser.
    Our method assigns a temperature to each diffusion process and allows processes to swap temperatures based on their relative performance.
    This dynamically reallocates promising, low-energy
    trajectories to colder, more concentrated sampling regimes while allowing higher-energy states to escape local minima through randomized exploration.
    Experiments on canonical graph-structured CO problems show that our approach consistently improves the quality of the best solution found, while only adding minimal computational overhead.
\end{abstract}

\section{Introduction}
\label{sec:introduction}

\Ac{CO} problems over graphs, such as the \ac{MIS}, \ac{MDS}, and maximum cut problems, are fundamental challenges in computer science and engineering.
Many of these problems are NP-hard and are commonly addressed using approximation algorithms or problem-specific heuristics.
Deep learning has emerged as an alternative paradigm, leveraging neural networks to learn problem-specific heuristics directly from data \citep{neural-co-with-rl, attention-learn-to-solve-routing-problems, erdos-goes-neural}.
Within this domain, generative models, and specifically diffusion models, have proven highly effective \citep{difusco}.
They frame \ac{CO} as sampling from a distribution over solutions and learn to iteratively denoise a uniform prior into high-quality \ac{CO} solutions.

At inference time, a common strategy for improving solution quality is to generate multiple candidates independently. Here, a pretrained denoiser generates $N$ candidate trajectories, and a non-learned decoder maps these candidates to feasible solutions. The final output is the best of the decoded solutions. Increasing $N$ can improve the best solution found, but requires proportionally more denoiser evaluations. The corresponding increase in wall-clock time depends on batching and available parallel hardware.
However, independent sampling at a fixed temperature does not use information about the relative progress of each trajectory in the ensemble to adapt each trajectory's sampling behavior. We therefore ask whether coordinating these trajectories can improve the best final solution under a fixed budget of denoiser evaluations.

\begin{figure}[t]
    \centering
    \includegraphics[width=\linewidth]{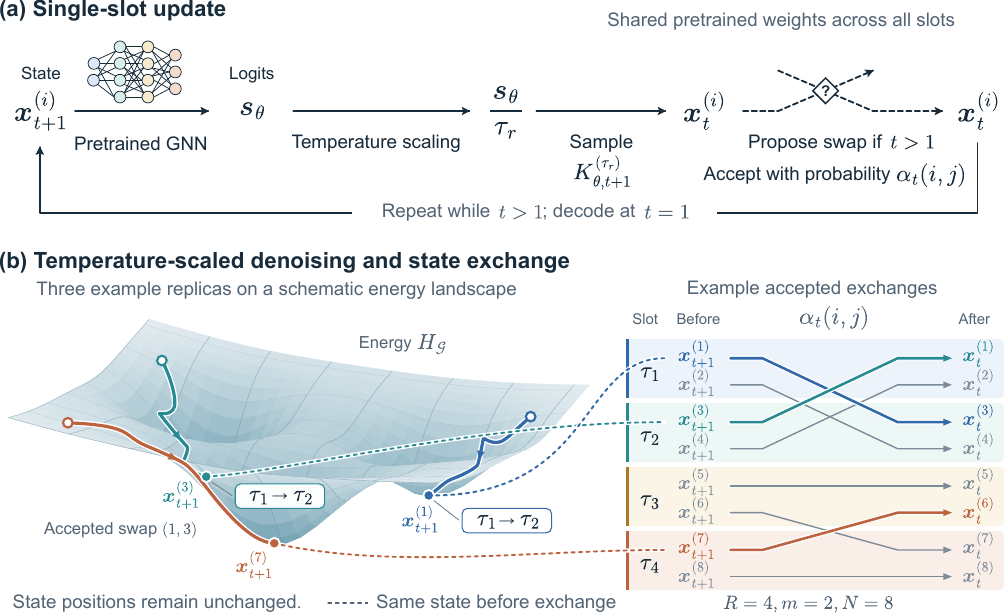}
    \caption{
        Overview of our approach.
        Top: One diffusion step of a single replica within an ensemble.
        We alternate between diffusion steps and replica swaps.
        Bottom: Swapping replicas between rungs on the temperature ladder.
        Neighboring temperature rungs are paired; in this example, the pairs are $(\tau_1, \tau_2), (\tau_3, \tau_4)$.
        A uniform one-to-one matching is sampled between the slots of paired temperatures.
        Swaps are proposed between matched slots and accepted with probability $\alpha_t(i, j)$.
    }
    \label{fig:overview}
\end{figure}

To address this limitation, we introduce \methodname{}, an inference-time procedure that coordinates concurrent denoising trajectories, which we call replicas, through parallel tempering. Our approach requires no retraining, fine-tuning, or architectural modifications to the underlying pretrained denoiser. We assign a temperature parameter to each replica slot, thereby controlling the level of randomness in its denoising steps.
Our approach periodically evaluates intermediate states and proposes state exchanges between slots at neighboring temperatures.
This mechanism favors assigning lower-energy states to colder, more concentrated sampling regimes and higher-energy states to hotter, more exploratory regimes, using intermediate energy as a surrogate for final decoded solution quality. \Cref{fig:overview} shows an overview of our approach.

Experiments on four graph-structured \ac{CO} problems (\ac{MIS}, \ac{MDS}, maximum clique, and maximum cut) show that our approach improves mean best-of-$N$ solution quality for both DiffUCO and SDDS, with small absolute runtime increases under the evaluated settings.
We make the following contributions:
\begin{itemize}[leftmargin=2em]
    \item \emph{A coordinated inference protocol}: We propose \methodname{}, a plug-and-play inference-time procedure that coordinates concurrent diffusion trajectories through state exchanges across temperatures without retraining the base model.
    \item \emph{Decoupled population and temperature ladder}: Since trajectories have only a limited number of denoising steps in which to move across the temperature ladder, we maintain multiple trajectories per temperature rung. This allows increasing the population size without increasing the number of neighboring exchanges needed to traverse the ladder.
    \item \emph{Empirical evaluation under a fixed denoiser budget}: We evaluate our approach across four graph-structured \ac{CO} problems using two pretrained diffusion backbones, demonstrating improvements in mean solution quality and measuring the accompanying inference-time overhead.
\end{itemize}

\section{Related work}
\label{sec:related-work}

\textbf{Neural combinatorial optimization.~}
Early neural approaches learn constructive heuristics through reinforcement learning, using
recurrent networks \citep{neural-co-with-rl},
graph embeddings \citep{learning-co-algorithms-over-graphs},
attention-based architectures \citep{attention-learn-to-solve-routing-problems}, or
progressive decision-deferring schemes \citep{learning-what-to-defer}.
Beyond reinforcement learning, Erd{\H{o}}s Goes Neural \citep{erdos-goes-neural, egn-anneal} combines unsupervised optimization of probabilistic relaxations with derandomized decoding, while Let the Flows Tell \citep{let-the-flows-tell} trains conditional GFlowNets to generate high-quality graph solutions.
\citet{intel} use a learned model to guide a parallel tree search, a paradigm later critically evaluated by \citet{dgl}.
These approaches motivate learned solution generation; our focus is on
coordinating trajectories from an existing diffusion solver at inference time.

\textbf{Diffusion for combinatorial optimization.~}
DIFUSCO \citep{difusco} introduced graph-based diffusion solvers, while
DiffUCO \citep{diffuco} and SDDS \citep{sdds} developed methods for training
discrete diffusion samplers against energy-based objectives without solution
labels. DISCO \citep{disco} improves solution quality and inference efficiency
through a residual-based diffusion formulation. Closely related,
GenSCO \citep{generation-as-search-operator} treats generation as a search
operator, alternating solution disruption and learned generation; its search
loop is supported by a dedicated solution-enhancement training procedure.
Our approach instead couples concurrent denoising trajectories through
temperature exchanges, using an existing pretrained denoiser without
retraining or fine-tuning.

\textbf{Inference-time steering and interacting particles.~}
Particle Guidance \citep{particle-guidance} couples diffusion trajectories through a joint potential to encourage diversity. Feynman--Kac steering \citep{fk-steering} resamples particles according to intermediate reward potentials, while soft value-based decoding \citep{svdd} guides generation using estimates of future reward. Related sequential Monte Carlo methods derive importance-sampling corrections and proposal mechanisms to control discrete diffusion models \citep{discrete-smc,discrete-smc-scaling}.
Our method uses energy-dependent exchanges between temperature groups:
each exchange preserves the current collection of states while changing
the sampling temperatures governing their subsequent evolution.

\textbf{Parallel tempering and generative sampling.~}
Parallel tempering coordinates replicas through exchanges across a temperature ladder \citep{replica-exchange-1,replica-exchange-2,replica-exchange-pairs}. Ensemble-based implementations and adaptive ladder selection were studied by \citet{vousden}. Accelerated Parallel Tempering \citep{apt} incorporates neural transports, including flows and diffusion processes, into parallel tempering while preserving asymptotic consistency under its stated assumptions.
Source Parallel Tempering \citep{source-parallel-tempering} uses parallel tempering to guide flow-based models, but operates on continuous latent priors in the source space before generation.
In contrast, our method couples discrete denoising trajectories across diffusion timesteps.
CREPE \citep{crepe} applies parallel tempering to inference-time control of pretrained diffusion models, including discrete masked diffusion, using replicas at different diffusion times. Our approach exchanges states at the same denoising time across different logit temperatures and assigns multiple replicas to each rung, allowing population size to increase without extending the ladder. We use this mechanism for finite-budget optimization. Temperature-scaled denoising generally does not preserve Boltzmann distributions, so the overall procedure does not inherit the classical parallel-tempering sampling guarantees.

\section{Background}
\label{sec:background}

\textbf{\Acl{CO}.~}
A \acf{CO} problem asks us to minimize an objective function $f: \mathcal{F} \rightarrow \mathbb{R}$ on a discrete set $\mathcal{F}$.
Without loss of generality, maximization problems can be converted to minimization problems by changing the sign.
Many \ac{CO} problems admit natural graph representations. In particular, much of the machine learning literature for \ac{CO} focuses on node subset problems \citep{erdos-goes-neural, let-the-flows-tell}.
Given a graph $\mathcal{G} = (V, E)$, the goal is to minimize an objective function $f_\mathcal{G}: \mathcal{F}_\mathcal{G} \rightarrow \mathbb{R}$ over subsets of nodes, where $\mathcal{F}_\mathcal{G} \subseteq \{0, 1\}^{|V|}$.
A relaxed energy function $H_\mathcal{G}: [0, 1]^{|V|} \rightarrow \mathbb{R}$ is often used to evaluate approximate continuous solutions or solutions that do not meet the constraints.
It usually consists of a term that relaxes the discrete objective function and one that softly enforces the \ac{CO} problem's constraints, and is designed to match the objective function $f_\mathcal{G}$ on $\mathcal{F}_\mathcal{G}$ \citep{erdos-goes-neural}.

Examples for frequently studied \ac{CO} problems are as follows.
The \emph{\acf{MIS}} problem asks us to find the largest subset of nodes $S \subseteq V$ in which no pair of nodes are neighbors.
In the \emph{\acf{MDS}} problem, the goal is to find the smallest subset $S \subseteq V$ such that each node in $V$ is either in $S$ or has a neighbor in $S$.
The \emph{maximum cut} problem asks us to find a subset $S \subseteq V$ that maximizes the number of edges between $S$ and $V \setminus S$.
Finally, there is the \emph{maximum clique} problem, where the goal is to find the largest clique, i.e.\@ a subset of nodes where each node is connected to each other node in the set.

\textbf{Diffusion for \ac{CO}.~}
Diffusion models for \ac{CO} frame solving \ac{CO} problems over graphs as generative modeling over the solution space $\mathcal{X} = \{0, \dots, k-1\}^{|V|}$ disregarding constraints, where $k = 2$ for node subset problems \citep{difusco}.
The diffusion model can be trained to approximately sample from a Boltzmann distribution based on the \ac{CO} problem's energy function $H_\mathcal{G}$:
\begin{equation*}
     p_\mathcal{G}^*(\bm{x})
    = \frac{1}{Z(\mathcal{G})}
    \exp \bigl(
        -\beta_B H_\mathcal{G}(\bm{x})
    \bigr),
    \qquad Z(\mathcal{G})
    = \sum_{\bm{x} \in \mathcal{X}}
    \exp \bigl(
        - \beta_B H_\mathcal{G}(\bm{x})
    \bigr),   
\end{equation*}
where $\beta_B > 0$ is an inverse temperature \citep{diffuco, sdds}.
This assigns the highest probabilities to solutions $\bm{x}$ with low energy.

Discrete diffusion connects the target distribution $p_\mathcal{G}^*$ to a simple reference distribution through a fixed corruption process and a learned denoising process.
The \emph{forward process} $q(\bm{x}_{1:T} \mid \bm{x}_0)$ is a non-trainable Markov chain that progressively corrupts an initial assignment $\bm{x}_0$.
For discrete variables, each variable is resampled from a categorical transition distribution determined by its current value and the timestep.
A sampled value may coincide with the current value or replace it with another of the $k$ possible categories.
The noise schedule progressively removes information about $\bm{x}_0$, making the terminal distribution approximately a factorized uniform prior,
\begin{equation*}
      p(\bm{x}_T)
    = \prod_{v \in V}
    \operatorname{Cat}\!\left(x_{T,v}; \frac{1}{k}\bm{1}\right).  
\end{equation*}
The \emph{reverse process} learns to undo this corruption.
Starting from an assignment sampled from $p(\bm{x}_T)$, it applies learned categorical transitions conditioned on the graph $\mathcal{G}$:
\begin{equation}
       p_\theta(\bm{x}_{0:T}\mid\mathcal{G})
    = p(\bm{x}_T)
    \prod_{t=1}^T
    p_\theta(\bm{x}_{t-1}\mid\bm{x}_t,\mathcal{G}). 
\end{equation}
The parameters $\theta$ are trained so that the resulting marginal distribution $p_\theta(\bm{x}_0\mid\mathcal{G})$ approximates the target distribution $p_\mathcal{G}^*$ over high-quality \ac{CO} solutions.

At inference time, the forward process is discarded entirely.
Generation proceeds by sampling initial states from the prior $p(\bm{x}_T)$ and sequentially evaluating the transitions $p_\theta(\bm{x}_{t-1} \mid \bm{x}_t)$ down to $t=0$.
The reverse transition can be parameterized via a \ac{GNN} $\bm{s}_\theta(\bm{x}_t, t, \mathcal{G}) \in \mathbb{R}^{|V| \times k}$, which predicts node-level logits based on the intermediate state $\bm{x}_t$, timestep $t$, and problem graph $\mathcal{G}$:
\begin{equation*}
        K_{\theta,t} (
        \bm{x}_{t-1} \mid \bm{x}_t,\mathcal{G}
    )
    = \prod_{v\in V} \Cat \Bigl(
        \bm{x}_{t-1,v}; \;
        \softmax \bigl(
            \bm{s}_{\theta,v} (\bm{x}_t,t,\mathcal{G})
        \bigr)
    \Bigr).
\end{equation*}
While the output of the reverse process $\bm{x}_0 \in \mathcal{X}$ is discrete, it is not guaranteed to meet the \ac{CO} problem's constraints.
To address this, the logits at the final denoising step $\bm{s}_\theta(\bm{x}_1, 1, \mathcal{G})$ are fed into a \emph{decoder} $\bm{d}_\mathcal{G}: \mathbb{R}^{|V| \times k} \rightarrow \mathcal{F}_\mathcal{G}$, an algorithm that constructs a feasible solution to the \ac{CO} problem.

\textbf{Parallel tempering.~}
Parallel tempering, also known as replica exchange \ac{MCMC} sampling \citep{replica-exchange-1, replica-exchange-2}, is an algorithm used in statistical physics and computational chemistry to sample from multimodal distributions $p^*(\bm{x}) \propto \exp(- H(\bm{x}) / \tau)$, where $\tau \in \mathbb{R}$ is a temperature and $H$ is an energy function.

Parallel tempering maintains a set of $N$ replicas operating at distinct temperatures along a temperature ladder $\mathcal{T} = \{\tau_1 < \tau_2 < \dots < \tau_N\}$, where typically $\tau_1 = 1$.
Each replica $i \in \{1, \dots, N\}$ evolves its sample $\bm{x}^{(i)}$ independently, e.g.\@ according to a standard \ac{MCMC} algorithm.
Replicas only interact periodically at predetermined intervals by swapping temperatures:
For each pair of neighboring temperatures $\tau_i, \tau_{i+1}$, replicas $i, \ j = i+1$ swap according to the Metropolis--Hastings criterion with probability
\begin{equation}
    \label{eq:mh-criterion}
    \alpha(i, j)
    = \min \Bigl\{
        1, \ \exp \bigl(
            (\beta_i - \beta_j)
            (H(\bm{x}^{(i)}) - H(\bm{x}^{(j)}))
        \bigr)
    \Bigr\},
\end{equation}
where $\beta_i = 1 / \tau_i$.
High-temperature replicas explore broad regions of state space and can cross high-energy barriers easily, while low-temperature replicas exploit local minima.

Conflicts can arise if swaps between replica pairs $(i, i+1)$ and $(i+1, i+2)$ are proposed in the same iteration.
To address this, a common approach is to group replica pairs $(i, i+1)$ into pairs with even $i$ and pairs with odd $i$, and to alternate between proposing swaps using even and odd pairs \citep{replica-exchange-pairs}.

\section{Method}
\label{sec:method}

We consider a \ac{CO} problem on a graph $\mathcal{G}$, with feasible solutions $\mathcal{F}_{\mathcal{G}} \subseteq \mathcal{X}$ and objective $f_{\mathcal{G}}: \mathcal{F}_{\mathcal{G}} \rightarrow \mathbb{R}$ to be minimized.
A pretrained discrete denoiser and a non-learned decoder together define a stochastic procedure for generating feasible solutions.
A common inference strategy \citep{difusco, diffuco, sdds} is to generate $N$ candidate solutions $\bm{x}^{(1)}, \dots, \bm{x}^{(N)} \in \mathcal{F}_\mathcal{G}$ independently and return
\[
    \bm{x}^* \in
    \argmin_{\bm{x} \in \{ \bm{x}^{(1)}, \dots, \bm{x}^{(N)} \}}
    f_{\mathcal{G}}(\bm{x}).
\]
Increasing $N$ can improve the best solution found, but also increases the computational cost.
We therefore ask whether coordinating a fixed number of denoising trajectories can improve their best final solution.

We propose \methodname{}, an inference-time procedure that couples concurrent denoising trajectories through temperature exchanges.
Different sampling temperatures control the randomness of the denoising transitions, while energy-dependent exchanges reassign intermediate states to these temperatures.
The procedure uses the same pretrained denoiser throughout and requires no retraining or fine-tuning.
\Cref{fig:overview} shows an overview of our approach.

Our objective is to improve
\begin{equation*}
        \mathbb{E} \left[
        \min_{i \in \{1, \dots, N\}}
        f_{\mathcal{G}} \bigl( \bm{x}^{(i)} \bigr)
    \right]
\end{equation*}
under a fixed inference budget.
We use parallel tempering as an optimization mechanism. Importantly, we do not assume that the resulting denoising trajectories 
are equilibrium samples from tempered Boltzmann distributions, as this is not required for the problem of finding high-quality solutions.

\paragraph{Temperature-scaled denoising.~} 
Let $\bm{s}_{\theta}(\bm{x}_t,t,\mathcal{G}) \in\mathbb{R}^{|V|\times k}$ denote the logits of a pretrained reverse transition.
For a sampling temperature $\tau>0$, define
\begin{equation}
    \label{eq:tempered-denoising}
    K_{\theta,t}^{(\tau)} (
        \bm{x}_{t-1} \mid \bm{x}_t,\mathcal{G}
    )
    = \prod_{v\in V} \Cat \left(
        \bm{x}_{t-1,v}; \;
        \softmax \left(
            \frac{1}{\tau} \bm{s}_{\theta,v} (\bm{x}_t,t,\mathcal{G})
        \right)
    \right).
\end{equation}
At $\tau=1$, this recovers the original reverse transition.
Larger temperatures flatten the categorical distributions, whereas smaller temperatures concentrate probability on the denoiser's preferred assignments.
Note that the sampling temperature $\tau$ used here is separate from the inverse temperature $\beta_B$ in the Boltzmann distribution used to train the model.

Temperature scaling acts on the transition distribution:
\begin{equation}
    \label{eq:transition-distribution}
    K_{\theta,t}^{(\tau)} (
        \bm{x}_{t-1} \mid \bm{x}_t,\mathcal{G}
    )
    \propto
    \left[
        K_{\theta,t}^{(1)} (
            \bm{x}_{t-1} \mid \bm{x}_t,\mathcal{G}
        )
    \right]^{1/\tau},
\end{equation}
where the normalizing constant depends on $\bm{x}$, $t$, and $\tau$.
Consequently, scaling the transition logits does not generally imply that intermediate or terminal states follow a Boltzmann distribution with temperature $\tau$.

\paragraph{Concurrent denoising with parallel tempering.~}
We maintain $N$ replica slots.
Each slot $i$ has an assigned temperature $\tau_i$ and contains
an intermediate state $\bm{x}_t^{(i)}$. Here, temperature subscripts index slots, and multiple slots may share the same temperature.
All replicas follow the same denoising schedule. Starting from independent samples of the original prior, each slot generates its next state using
\begin{equation}
    \label{eq:next-state-sampling}
    \bm{x}_t^{(i)}
    \sim
    K_{\theta,t+1}^{(\tau_i)}
    \bigl(
        \,\cdot\mid
        \bm{x}_{t+1}^{(i)},\mathcal{G}
    \bigr).
\end{equation}

After a denoising transition, we evaluate each state using the energy function $H_{\mathcal{G}}$.
This acts as a surrogate for the final solution quality at $t = 0$ after decoding, which we cannot measure directly.

For two slots $i$ and $j$ at neighboring temperatures, we propose exchanging their states and accept with probability
\begin{equation}
    \label{eq:denoising-exchange}
    \alpha_t(i,j)
    =
    \min\left\{
        1,\,
        \exp\left[
            (\beta_i-\beta_j)
            \left(
                H_\mathcal{G}(\bm{x}_t^{(i)})
                -
                H_\mathcal{G}(\bm{x}_t^{(j)})
            \right)
        \right]
    \right\},
    \qquad
    \beta_i=\frac{1}{\tau_i}.
\end{equation}
Temperatures remain attached to slots, so an accepted exchange changes the temperature used for the subsequent evolution of each exchanged state.
If $\tau_i<\tau_j$, an exchange that moves the lower-scoring, i.e.\@ more promising, state into slot $i$ is always accepted.
The reverse exchange may also be accepted, allowing states to move between more concentrated and more randomized denoising transitions.

An exchange preserves the current collection of states, including its best score. Its effect is therefore on subsequent denoising:
It changes the sampling temperature each state encounters.
This differs from resampling, which can duplicate some states and discard others \citep{fk-steering}.

\begin{algorithm}[t]
    \caption{Concurrent denoising with temperature exchange.}
    \label{alg:denoising-with-parallel-tempering}
    \begingroup
    \algtext*{EndFor}
    \algtext*{EndIf}
    \algrenewcommand\algorithmicindent{0.9em}
    \algrenewcommand\alglinenumber[1]{\scriptsize #1:}

    \noindent\textbf{Inputs:}
    Pretrained denoiser $\bm{s}_{\theta}$, graph $\mathcal{G}$,
    decoder $\bm{d}_{\mathcal{G}}$, prior $p_T$,
    temperature ladder $0<\tau_1<\cdots<\tau_R$,
    replicas per rung $m$, diffusion steps $T\geq2$,
    energy function $H_{\mathcal{G}}$, objective function $f_\mathcal{G}$.\par
    \noindent\textbf{Setup:}
    $N\gets Rm$; partition replica slots $\{1,\ldots,N\}$ into disjoint
    $\mathcal{I}_1,\ldots,\mathcal{I}_R$ with $|\mathcal{I}_r|=m$.
    Set $\beta_i\gets\tau_r^{-1}$ for each $r\in\{1,\ldots,R\}$
    and $i\in\mathcal{I}_r$.\par
    \smallskip

    \noindent
    \begin{minipage}[t]{0.45\linewidth}
        \textbf{(a) Denoising and selection}
        \begin{algorithmic}[1]
            \State $\bm{x}_T^{(i)}\overset{\mathrm{iid}}{\sim}p_T$, $i=1,\ldots,N$; $q\gets0$
            \For{$t=T-1,\ldots,1$}
                \For{$r=1,\ldots,R$, $i\in\mathcal{I}_r$}
                    \State $\bm{x}_t^{(i)}\sim K_{\theta,t+1}^{(\tau_r)}
                    (\cdot\mid\bm{x}_{t+1}^{(i)},\mathcal{G})$
                \EndFor
                \If{$t>1$}
                    \State $\bm{X}_t\gets(\bm{x}_t^{(i)})_{i=1}^{N}$
                    \State $\bm{X}_t\gets\Call{Exchange}{\bm{X}_t,q}$
                    \State $q\gets q+1$
                \EndIf
            \EndFor
            \For{$i=1,\ldots,N$}
                \State $\bm{x}^{(i)}\gets\bm{d}_{\mathcal{G}}\!\left(
                    \bm{s}_{\theta}(\bm{x}_1^{(i)},1,\mathcal{G})\right)$
            \EndFor
            \State $i^\star\in\argmin_{i\in\{1,\ldots,N\}}
                f_{\mathcal{G}}(\bm{x}^{(i)})$
            \State \Return $\bm{x}^{(i^\star)}$
        \end{algorithmic}
    \end{minipage}\hfill
    \begin{minipage}[t]{0.49\linewidth}
        \textbf{(b) \textsc{Exchange}$(\bm{X}_t,q)$}
        \begin{algorithmic}[1]
            \State $\mathcal{P}\gets\mathcal{P}^{\mathrm{odd}}$ if $q$ is even,
            \Statex \hspace{\algorithmicindent}otherwise $\mathcal{P}\gets\mathcal{P}^{\mathrm{even}}$
            \ForAll{$(r,s)\in\mathcal{P}$}
                \State $\mathcal{M}\gets\Call{UniformMatch}{\mathcal{I}_r,\mathcal{I}_s}$
                \ForAll{$(i,j)\in\mathcal{M}$}
                    \State $\Delta\gets(\beta_i-\beta_j)\bigl[
                        H_{\mathcal{G}}(\bm{x}_t^{(i)})-H_{\mathcal{G}}(\bm{x}_t^{(j)})\bigr]$
                    \State $u\sim\mathrm{Uniform}(0,1)$
                    \If{$\log u<\min\{0,\Delta\}$}
                        \State $(\bm{x}_t^{(i)},\bm{x}_t^{(j)})
                        \gets(\bm{x}_t^{(j)},\bm{x}_t^{(i)})$
                    \EndIf
                \EndFor
            \EndFor
            \State $\bm{X}_t=(\bm{x}_t^{(i)})_{i=1}^{N}$
            \State \Return $\bm{X}_t$
        \end{algorithmic}
    \end{minipage}
    \par\smallskip

    \textsc{UniformMatch} draws a uniformly random one-to-one matching.
    $\mathcal{P}^{\mathrm{odd}}$ and $\mathcal{P}^{\mathrm{even}}$
    are the rung pairings defined in Eq.~\eqref{eq:rung-odd-even}. Sampling: Eqs.~\eqref{eq:tempered-denoising},
    \eqref{eq:next-state-sampling}; exchange: Eq.~\eqref{eq:denoising-exchange}; decoding: Eq.~\eqref{eq:state-decoding}.\par
    \endgroup
\end{algorithm}

\paragraph{Decoupling the temperature ladder from population size.~} 
We use a ladder of $R$ distinct temperatures, $\mathcal{T} = \{ \tau_1 < \dots < \tau_R \}$, and assign $m$ replica slots to each temperature, giving $N = Rm$ replicas in total. Let $\mathcal{I}_r$ denote the slots assigned to rung $r$. In this grouped notation, temperature subscripts index ladder rungs rather than slots; for each $i \in \mathcal{I}_r$, the slot's inverse temperature is $\beta_i = 1/\tau_r$. This construction follows the ensemble-tempering principle, maintaining multiple states at each temperature \citep{vousden}.

This choice is motivated by the fact that using one replica per rung ties population size to the number of temperatures, potentially causing multiple problems.
Since a state can move at most one rung per exchange step, a longer ladder can limit movement between its endpoints within a short denoising horizon.
Allowing $R$ and $N$ to be adjusted separately allows us to increase the number of candidates without increasing the number of rungs they must traverse for the same temperature change.
Moreover, increasing the number of temperatures means either increasing the maximum temperature, or decreasing the distance between temperatures.
In the first case, larger parts at the top end of the temperature ladder turn to pure noise, ignoring the diffusion model's learned distribution.
In the second case, if the difference between neighboring temperatures becomes too small, the acceptance probability according to the Metropolis--Hastings criterion approaches 1, meaning that neighboring replicas will almost always swap regardless of energy.
These problems are resolved by assigning multiple replica slots to each temperature.

To avoid conflicting exchanges, we alternate between the disjoint rung pairings
\begin{equation}
\label{eq:rung-odd-even}
    \mathcal{P}^{\mathrm{odd}}
    = \{ (1, 2), (3, 4), \dots \},
    \qquad
    \mathcal{P}^{\mathrm{even}}
    = \{ (2, 3), (4, 5), \dots \},
\end{equation}
retaining only pairs with indices in $\{1,\ldots,R\}$.
For each active pair $(r,s)$, we sample a uniform one-to-one matching between $\mathcal{I}_r$ and $\mathcal{I}_s$ and apply \Cref{eq:denoising-exchange} to each matched pair.
Because pairs are disjoint, their exchanges can be evaluated in parallel.

\paragraph{Final decoding and computational cost.~} 
After reaching $t=1$, each replica is decoded using
\begin{equation}
    \label{eq:state-decoding}
    \bm{x}^{(i)}
    =
    \bm{d}_{\mathcal{G}}
    \left(
        \bm{s}_{\theta}
        (\bm{x}_1^{(i)},1,\mathcal{G})
    \right)
    \in\mathcal{F}_{\mathcal{G}},   
\end{equation}
where $\bm{d}_{\mathcal{G}}$ is the same feasibility-ensuring decoder used by the base diffusion model.
We return the candidate with the smallest $f_{\mathcal{G}}$.
With $T-1$ sampled transitions and one final denoiser evaluation for decoding, the procedure uses $NT$ per-replica denoiser evaluations, matching independent generation with the same $N$ and $T$.
The additional cost consists of intermediate score evaluations, random matching, and state exchanges.
Each exchange sweep proposes at most $N/2$ swaps.
An exchange immediately before a common deterministic decoder only permutes the final collection of candidates and cannot change the returned objective value.
It can therefore be omitted.

Exchanges assign intermediate states to different temperatures for subsequent denoising.
The acceptance rule favors moving lower-energy states to colder temperatures, where sampling is more concentrated.
Since lower energy corresponds to better objective values, this aims to refine promising states while allowing others to explore through more randomized transitions.
An exchange itself leaves the current candidates unchanged and its benefit depends on whether intermediate energy predicts final decoded quality and whether the temperature reassignment improves subsequent denoising.
Because temperature-scaled denoising does not generally preserve Boltzmann distributions, we treat parallel tempering as an optimization heuristic without claiming the sampling guarantees of classical parallel tempering.
We summarize the complete denoising procedure in \Cref{alg:denoising-with-parallel-tempering}.

\begin{table}[t]
	\centering
	\caption{
        Results on \ac{MIS}.
        Top: diffusion methods. Center: learned non-diffusion methods. Bottom: non-learned heuristics.
    }
    \label{tab:results-mis}
    \resizebox{\textwidth}{!}{%
	\begin{tabular}{lcccccc}
		\toprule
                                       & \multicolumn{3}{c}{RB Small} & \multicolumn{3}{c}{RB Large} \\
		\cmidrule(lr{1mm}){2-4}\cmidrule(lr{1mm}){5-7}
		Method                         & IS Size $\uparrow$    & Relative Gap $\downarrow$ & Time  &
                         IS Size $\uparrow$    & Relative Gap $\downarrow$ & Time  \\
        \midrule
		DiffUCO                        & \meanstd{19.74}{0.03} & \meanstd{1.79\%}{0.15}    & 12ms  &
                                         \meanstd{41.65}{0.03} & \meanstd{3.48\%}{0.08}    & 303ms \\
        DiffUCO + Temp. Ladder         & \meanstd{19.93}{0.04} & \meanstd{0.83\%}{0.18}    & 12ms  &
                                         \meanstd{41.66}{0.02} & \meanstd{3.45\%}{0.05}    & 303ms \\
		DiffUCO + \methodname{} (ours) & \meanstd{20.01}{0.01} & \meanstd{0.47\%}{0.03}    & 13ms  &
                                         \meanstd{41.68}{0.02} & \meanstd{3.40\%}{0.06}    & 303ms \\
		SDDS                           & \meanstd{19.95}{0.02} & \meanstd{0.72\%}{0.12}    & 12ms  &
                                         \meanstd{41.97}{0.07} & \meanstd{2.73\%}{0.16}    & 304ms \\
        SDDS + Temp. Ladder            & \meanstd{20.01}{0.02} & \meanstd{0.44\%}{0.12}    & 12ms  &
                                         \textbf{\meanstd{41.99}{0.02}} & \textbf{\meanstd{2.69\%}{0.09}}    & 303ms \\
		SDDS + \methodname{} (ours)    & \textbf{\meanstd{20.05}{0.02}} & \textbf{\meanstd{0.26\%}{0.08}} & 13ms  &
                                         \textbf{\meanstd{42.01}{0.03}} & \textbf{\meanstd{2.64\%}{0.08}} & 304ms \\
		\midrule
        LwtD           & 19.01 & 5.42\%  & 154ms  & 32.32 & 25.10\% & 906ms  \\
        INTEL          & 18.47 & 8.11\%  & 1.57s  & 34.47 & 20.12\% & 2.43s  \\
        DGL            & 17.36 & 13.61\% & 1.53s  & 34.50 & 20.05\% & 2.87s  \\
        LTFT           & 19.18 & 4.57\%  & 64ms   & 37.48 & 13.14\% & 524ms  \\
		\midrule
		KaMIS          & 20.10 & ---     & 10.10s & 43.15 & ---     & 14.83s \\
		Gurobi         & 19.98 & 0.60\%  & 5.71s  & 40.90 & 5.21\%  & 15.65s \\
		\bottomrule
	\end{tabular}
    }
\end{table}

\begin{table}[t]
	\centering
	\caption{
        Results on \ac{MDS}.
        Top: diffusion methods. Center: learned non-diffusion methods. Bottom: non-learned heuristics.
    }
    \label{tab:results-mds}
    \resizebox{\textwidth}{!}{%
	\begin{tabular}{lcccccc}
		\toprule
                                       & \multicolumn{3}{c}{BA Small} & \multicolumn{3}{c}{BA Large} \\
		\cmidrule(lr{1mm}){2-4}\cmidrule(lr{1mm}){5-7}
		Method                         & DS Size $\downarrow$   & Relative Gap $\downarrow$ & Time  &
                                         DS Size $\downarrow$   & Relative Gap $\downarrow$ & Time  \\
		\midrule
		DiffUCO                        & \meanstd{28.01}{0.05}  & \meanstd{0.43\%}{0.18}    & 10ms  &
                                         \meanstd{104.01}{0.02} & \meanstd{0.00\%}{0.02}    & 56ms  \\
		DiffUCO + \methodname{} (ours) & \meanstd{27.91}{0.01}  & \meanstd{0.08\%}{0.04}    & 11ms  &
                                         \meanstd{103.92}{0.01} & \meanstd{-0.08\%}{0.01}   & 56ms  \\
		SDDS                           & \meanstd{27.93}{0.01}  & \meanstd{0.14\%}{0.04}    & 10ms  &
                                         \meanstd{103.90}{0.02} & \meanstd{-0.11\%}{0.02}   & 56ms  \\
		SDDS + \methodname{} (ours)    & \textbf{\meanstd{27.90}{0.00}}  & \textbf{\meanstd{0.02\%}{0.01}}  & 11ms  &
                                         \textbf{\meanstd{103.84}{0.03}} & \textbf{\meanstd{-0.16\%}{0.03}} & 57ms  \\
		\midrule
        EGN            & 30.68 & 10.00\% & 120ms  & 116.76 & 12.26\% & 472ms  \\
        EGN-Anneal     & 29.24 & 4.84\%  & 122ms  & 111.50 & 7.20\%  & 470ms  \\
        LTFT           & 28.61 & 2.58\%  & 280ms  & 110.28 & 6.03\%  & 3.86s  \\
        \midrule
		Gurobi         & 27.89 & ---     & 214ms  & 104.01 & ---     & 1.66s  \\
		\bottomrule
	\end{tabular}
    }
\end{table}

\begin{table}[t]
	\centering
	\caption{
        Results on maximum cut.
        Top: diffusion methods. Center: learned non-diffusion methods. Bottom: non-learned heuristics.
    }
    \label{tab:results-max-cut}
    \resizebox{\textwidth}{!}{%
	\begin{tabular}{lcccccc}
		\toprule
                                       & \multicolumn{3}{c}{BA Small}     & \multicolumn{3}{c}{BA Large}     \\
		\cmidrule(lr{1mm}){2-4}\cmidrule(lr{1mm}){5-7}
		Method                         & Cut Size $\uparrow$     & Relative Gap $\downarrow$ & Time  &
                                         Cut Size $\uparrow$     & Relative Gap $\downarrow$ & Time  \\
		\midrule
		DiffUCO                        & \meanstd{734.18}{0.08}  & \meanstd{-0.45\%}{0.01}   &  9ms  &
                                         \meanstd{2966.36}{0.17} & \meanstd{-0.75\%}{0.01}   & 37ms  \\
		DiffUCO + \methodname{} (ours) & \textbf{\meanstd{734.55}{0.01}}  & \textbf{\meanstd{-0.50\%}{0.00}} & 10ms  &
                                         \textbf{\meanstd{2966.64}{0.18}} & \textbf{\meanstd{-0.76\%}{0.00}} & 38ms  \\
		SDDS                           & \meanstd{734.29}{0.10}  & \meanstd{-0.47\%}{0.01}  &  9ms  &
                                         \meanstd{2966.29}{0.26} & \meanstd{-0.74\%}{0.01}  & 38ms  \\
		SDDS + \methodname{} (ours)    & \textbf{\meanstd{734.56}{0.05}}  & \textbf{\meanstd{-0.50\%}{0.01}} & 10ms  &
                                         \textbf{\meanstd{2966.89}{0.30}} & \textbf{\meanstd{-0.76\%}{0.01}} & 38ms  \\
		\midrule
        EGN            & 693.45 & 5.12\%  & 92ms   & 2870.34 & 2.51\%  & 338ms  \\
        EGN-Anneal     & 696.73 & 4.67\%  & 90ms   & 2863.23 & 2.76\%  & 336ms  \\
        LTFT           & 704.30 & 3.64\%  & 354ms  & 2864.61 & 2.71\%  & 2.56s  \\
        \midrule
		Gurobi         & 730.87 & ---     & 1.57s  & 2944.38 & ---     & 7.86s  \\
		\bottomrule
	\end{tabular}
    }
\end{table}

\section{Experiments}
\label{sec:experiments}

We evaluate whether coordinating diffusion trajectories through temperature exchange improves solution quality under a fixed budget of denoiser evaluations.
Our experiments cover four graph \ac{CO} problems in \ac{MIS}, \ac{MDS}, maximum cut, and maximum clique and compare \methodname{} with independent sampling using the same diffusion backbones, alongside established \ac{CO} baselines.
We report both solution quality and inference time to assess the gains from coordination and its computational overhead.
We build upon DiffUCO \citep{diffuco} and SDDS \citep{sdds}, using the same model architectures and trained weights. For \methodname{}, we use $N = 100$ replicas and 18 diffusion steps in our experiments. To allow for a fair comparison, all diffusion baselines also sample 100 solutions using the same number of diffusion steps, and the best result is reported.
We use conditional expectation \citep{diffuco} as the decoder for \methodname{} and all diffusion baselines.

\textbf{\ac{CO} problems and datasets.~}
Our evaluation closely follows the protocol established by \citet{let-the-flows-tell, diffuco, sdds}.
We evaluate on four \ac{CO} problems: \ac{MIS}, \ac{MDS}, maximum cut, and maximum clique.
We use RB graphs \citep{rb-graphs} for \ac{MIS} and max clique, and \ac{BA} graphs \citep{ba-graphs} for \ac{MDS} and max cut.
For each dataset, we evaluate on a small variant with 200-300 nodes and a large variant with 800-1200 nodes.
For max clique, we only evaluate on RB small, following \citet{diffuco, sdds}.
All test sets contain 1,000 instances.
The energy functions used for each \ac{CO} problem are detailed in \Cref{sec:energy-functions}.

\begin{wraptable}[14]{R}{0.62\textwidth}
    \vspace{-4.5ex}
    \centering
    \caption{
        Results on maximum clique.
        Top: diffusion methods. Center: learned non-diffusion methods. Bottom: non-learned heuristics.
    }
    \label{tab:results-max-clique}
    \resizebox{\linewidth}{!}{%
    \begin{tabular}{lccc}
        \toprule
                                       & \multicolumn{3}{c}{RB Small} \\
        \cmidrule(lr{1mm}){2-4}
        Method                         & Clique Size $\uparrow$    & Relative Gap $\downarrow$ & Time  \\
        \midrule
        DiffUCO                        & \meanstd{18.24}{0.14} & \meanstd{4.27\%}{0.74}    & 23ms  \\
        DiffUCO + \methodname{} (ours) & \meanstd{18.91}{0.13} & \meanstd{0.75\%}{0.67}    & 24ms  \\
        SDDS                           & \meanstd{18.98}{0.00} & \meanstd{0.37\%}{0.02}    & 23ms  \\
        SDDS + \methodname{} (ours)    & \textbf{\meanstd{18.99}{0.00}} & \textbf{\meanstd{0.33\%}{0.03}}    & 24ms  \\
        \midrule
        EGN                            & 12.02 & 36.90\%  & 82ms  \\
        EGN-Anneal                     & 14.10 & 25.98\%  & 82ms  \\
        LTFT                           & 16.24 & 14.75\%  & 84ms  \\
        \midrule
        Gurobi                         & 19.05 & ---      & 230ms \\
        \bottomrule
    \end{tabular}
    }
\end{wraptable}

\textbf{Baselines.~}
Our primary diffusion baselines are SDDS \citep{sdds} and DiffUCO
{\defcitealias{diffuco}{Sano\-kowski et al., 2024} \citepalias{diffuco}}.
On \ac{MIS}, we compare against learned unsupervised baselines \emph{let the flows tell} (LTFT) \citep{let-the-flows-tell} and \emph{learning what to defer} (LwtD) \citep{learning-what-to-defer}, supervised baselines DGL \citep{dgl} and INTEL \citep{intel}, as well as the non-learned mixed-integer program solver Gurobi \citep{gurobi} and the non-learned heuristic solver KaMIS \citep{kamis}.
On \ac{MDS}, max cut, and max clique, we compare with learned unsupervised baselines LTFT, \emph{Erd\H{o}s goes neural} (EGN) \citep{erdos-goes-neural}, and an annealed variant EGN-Anneal \citep{egn-anneal}, as well as Gurobi.
In our tables, results for all non-diffusion baselines are reported as in \citep{let-the-flows-tell}.
We note that Gurobi and KaMIS often have substantially longer runtimes than the other methods and should therefore not be compared directly in those cases.

\textbf{Metrics.~}
We report the value of the objective function $f_\mathcal{G}$, i.e.\@ the size of the solution set (for \ac{MIS}, \ac{MDS}, and max clique) or resulting cut (for max cut). Mean and standard deviation are calculated over three published reference models\footnote{See \Cref{sec:hyperparameters} for details} trained with different seeds.
We also measure the execution time, averaged per \ac{CO} problem instance.
Finally, we report the relative gap to the best non-learned baseline.
The relative gap (in \%) is calculated as
$(f_\mathcal{G}(\bm{x}) - f_\mathcal{G}(\bm{x}^*)) / f_\mathcal{G}(\bm{x}^*) \cdot 100$
for minimization problems and
$(f_\mathcal{G}(\bm{x}^*) - f_\mathcal{G}(\bm{x})) / f_\mathcal{G}(\bm{x}^*) \cdot 100$
for maximization problems.
Here, $\bm{x}$ is the learned method's predicted solution and $\bm{x}^*$ is the solution from the best non-learned baseline.
Note that $\bm{x}^*$ is not necessarily the optimal solution, and the gap becomes negative if $\bm{x}$ is better than $\bm{x}^*$.

\textbf{Results.~}
We summarize our main empirical results in \Cref{tab:results-mis,tab:results-mds,tab:results-max-cut,tab:results-max-clique}.
The best result among learned methods as well as each result for which the respective mean lies within one standard deviation of the best method are marked in bold.
Across all four \ac{CO} problems and on both small and large graphs, incorporating \methodname{} consistently improves the solution quality of the base diffusion models, DiffUCO and SDDS.
SDDS + \methodname{} shows the best mean solution quality on all \ac{CO} problems and datasets among evaluated learned methods.

On \ac{MIS} (\Cref{tab:results-mis}), \methodname{} substantially reduces the relative gap to the best non-learned baseline.
For instance, applying our method to DiffUCO reduces the relative gap from $1.79\%$ to $0.47\%$ on RB small, while SDDS + \methodname{} achieves the best overall diffusion performance with a relative gap of $0.26\%$.
\begin{wrapfigure}[19]{R}{0.48\textwidth}
    \centering
    \includegraphics[width=\linewidth]{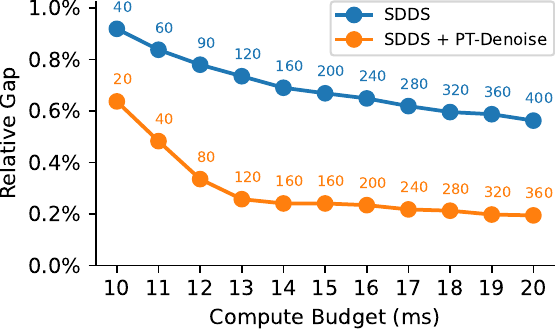}
    \vspace{-1ex}
    \caption{
        The best relative gap achievable on MIS, RB small within each given per-instance compute budget. The numbers next to each data point indicate how many replicas can be processed within the compute budget.
    }
    \label{fig:gap-per-compute-budget}
\end{wrapfigure}
The improvement persists on RB large, though to a much smaller extent.
In addition to this, the augmented diffusion models substantially outperform all learned non-diffusion baselines and outperform Gurobi, approaching the performance of KaMIS while being orders of magnitude faster.

As an ablation, we also ran DiffUCO and SDDS on MIS using only a ladder of different temperatures, but without temperature swaps.
For both diffusion models and both problem sizes, this performs better than a single temperature but worse than \methodname{}.
For \ac{MDS} (\Cref{tab:results-mds}), augmenting the base models with \methodname{} strictly improves the dominating set sizes.
Similarly, on max cut (\Cref{tab:results-max-cut}), \methodname{} increases the cut size across the board, but to a much lesser extent.
On max clique (\Cref{tab:results-max-clique}), where we only evaluate on small graphs, applying our method to DiffUCO reduces the relative gap from 4.27\% to 0.75\%, and applying it to SDDS results in the best performance overall by a small margin.
In all of our experiments, using \methodname{} adds at most 1ms per \ac{CO} problem instance to the inference time.
This means that any solution quality gained by using our method comes at very little additional computational cost.

\Cref{fig:gap-per-compute-budget} visualizes the solution quality achievable within a given per-instance time budget. This only uses replica counts divisible by 10, since we use 10 temperatures for \methodname{}. Relative gaps w.r.t.\@ KaMIS and inference times are averaged over three trained models. At every compute budget, using \methodname{} reduces the relative gap.

\section{Conclusion}
\label{sec:conclusion}

We introduced \methodname{}, a novel inference-time procedure that enhances discrete diffusion models for graph-structured \ac{CO} by coupling concurrent denoising trajectories through parallel tempering.
By assigning a temperature to each diffusion process and allowing them to swap temperatures, our approach enables both exploration and local refinement without requiring retraining or fine-tuning of the base model.
Our empirical evaluations across four canonical \ac{CO} problems demonstrate that our approach consistently improves the solution quality of state-of-the-art diffusion models, while incurring minimal computational overhead.

\textbf{Limitations and future work.~}
While this work focuses explicitly on \ac{CO}, the parallel tempering approach is general and could be applied to other diffusion domains.
Consequently, adapting \methodname{} to other application areas remains an open direction for future research.
Our approach relies on a relaxed energy function $H_\mathcal{G}$ to evaluate particles and guide temperature swaps.
While $H_\mathcal{G}$ serves as an effective heuristic surrogate, it may not perfectly correlate with the final decoded solution quality at early denoising timesteps.
A potential solution could lie in training a lightweight neural network to predict the final solution quality based on intermediate states.

\clearpage

\subsection*{AI use statement}


In this work, we used \acp{LLM} to help implement our method, write the scripts used to generate plots visualizing experimental results, and to suggest and improve formulations in the manuscript.
\acp{LLM} were not used to develop theoretical models, conceptual frameworks or mathematical claims, to propose or refine hypotheses, to design or provide feedback on research  methodology or experiments, to support qualitative and thematic data analysis, or to interpret results.
All synthetic datasets were created using non-learned algorithms, with no \ac{LLM}-assisted cleaning or reformatting.
The remaining required disclosure tasks---assisting in the writing of proofs, providing critical ingredients for proving mathematical claims, and assisting with translation---are not applicable to this work.
All \ac{LLM}-generated code and text has been manually reviewed by the authors.
We take responsibility for the final content of this work, including the code and text produced with the aid of \acp{LLM}.

\subsection*{Reproducibility statement}


The code required to reproduce the experiments was submitted as supplementary material, and will be made public once the paper is accepted.
This includes scripts for re-generating the datasets.
We list hyperparameters and checkpoints used in the experiments in \Cref{sec:hyperparameters}, and the hardware we used in \Cref{sec:hardware}.

\bibliography{references}

@article{neural-co-with-rl,
    title = {Neural combinatorial optimization with reinforcement learning},
    author = {Bello, Irwan and Pham, Hieu and Le, Quoc V and Norouzi, Mohammad and Bengio, Samy},
    journal = {arXiv preprint arXiv:1611.09940},
    year = {2016}
}

@inproceedings{learning-co-algorithms-over-graphs,
    title = {Learning Combinatorial Optimization Algorithms over Graphs},
    author = {Khalil, Elias and Dai, Hanjun and Zhang, Yuyu and Dilkina, Bistra and Song, Le},
    booktitle = {Advances in Neural Information Processing Systems},
    pages = {},
    publisher = {Curran Associates, Inc.},
    volume = {30},
    year = {2017}
}

@inproceedings{attention-learn-to-solve-routing-problems,
    title = {Attention, Learn to Solve Routing Problems!},
    author = {Wouter Kool and Herke van Hoof and Max Welling},
    booktitle = {International Conference on Learning Representations},
    year = {2019},
}

@inproceedings{erdos-goes-neural,
    author = {Karalias, Nikolaos and Loukas, Andreas},
    booktitle = {Advances in Neural Information Processing Systems},
    pages = {6659--6672},
    publisher = {Curran Associates, Inc.},
    title = {Erd{\H{o}}s Goes Neural: {An} Unsupervised Learning Framework for Combinatorial Optimization on Graphs},
    volume = {33},
    year = {2020}
}

@inproceedings{let-the-flows-tell,
    title = {Let the Flows Tell:  Solving Graph Combinatorial Problems with {GFlowNets}},
    author = {Zhang, Dinghuai and Dai, Hanjun and Malkin, Nikolay and Courville, Aaron C and Bengio, Yoshua and Pan, Ling},
    booktitle = {Advances in Neural Information Processing Systems},
    pages = {11952--11969},
    publisher = {Curran Associates, Inc.},
    volume = {36},
    year = {2023}
}

@InProceedings{learning-what-to-defer,
    title = {Learning What to Defer for Maximum Independent Sets},
    author = {Ahn, Sungsoo and Seo, Younggyo and Shin, Jinwoo},
    booktitle = {Proceedings of the 37th International Conference on Machine Learning},
    pages = {134--144},
    year = {2020},
    volume = {119},
    series = {Proceedings of Machine Learning Research},
    month = {Jul},
    day = {13--18},
    publisher = {PMLR},
}

@inproceedings{intel,
    title = {Combinatorial Optimization with Graph Convolutional Networks and Guided Tree Search},
    author = {Li, Zhuwen and Chen, Qifeng and Koltun, Vladlen},
    booktitle = {Advances in Neural Information Processing Systems},
    publisher = {Curran Associates, Inc.},
    volume = {31},
    year = {2018},
}

@inproceedings{dgl,
    title = {What's wrong with deep learning in tree search for combinatorial optimization},
    author = {B{\"o}ther, Maximilian and Ki{\ss}ig, Otto and Taraz, Martin and Cohen, Sarel and Seidel, Karen and Friedrich, Tobias},
    booktitle = {International Conference on Learning Representations},
    year = {2022},
}

@article{egn-anneal,
    title = {Annealed training for combinatorial optimization on graphs},
    author = {Sun, Haoran and Guha, Etash K and Dai, Hanjun},
    journal = {arXiv preprint arXiv:2207.11542},
    year = {2022},
}

@inproceedings{difusco,
    title = {{DIFUSCO}: Graph-based Diffusion Solvers for Combinatorial Optimization},
    author = {Sun, Zhiqing and Yang, Yiming},
    booktitle = {Advances in Neural Information Processing Systems},
    pages = {3706--3731},
    publisher = {Curran Associates, Inc.},
    volume = {36},
    year = {2023}
}

@InProceedings{diffuco,
    title = {A Diffusion Model Framework for Unsupervised Neural Combinatorial Optimization},
    author = {Sanokowski, Sebastian and Hochreiter, Sepp and Lehner, Sebastian},
    booktitle = {Proceedings of the 41st International Conference on Machine Learning},
    pages = {43346--43367},
    year = {2024},
    volume = {235},
    series = {Proceedings of Machine Learning Research},
    month = {Jul},
    day = {21--27},
    publisher = {PMLR},
}

@inproceedings{sdds,
    title = {Scalable Discrete Diffusion Samplers: Combinatorial Optimization and Statistical Physics},
    author = {Sanokowski, Sebastian and Berghammer, Wilhelm and Wang, Haoyu and Ennemoser, Martin and Hochreiter, Sepp and Lehner, Sebastian},
    booktitle = {International Conference on Learning Representations},
    pages = {87053--87082},
    year = {2025},
}

@article{disco,
    title = {{DISCO}: Efficient Diffusion Solver for large-scale Combinatorial Optimization problems},
    author = {Hang Zhao and Kexiong Yu and Yuhang Huang and Renjiao Yi and Chenyang Zhu and Kai Xu},
    journal = {Graphical Models},
    volume = {141},
    year = {2025},
}

@inproceedings{generation-as-search-operator,
    title = {Generation as Search Operator for Test-Time Scaling of Diffusion-based Combinatorial Optimization},
    author = {Li, Yang and Chen, Lvda and Wang, Haonan and Wang, Runzhong and Yan, Junchi},
    booktitle = {Advances in Neural Information Processing Systems},
    pages = {127168--127196},
    publisher = {Curran Associates, Inc.},
    volume = {38, Main Conference},
    year = {2025}
}

@inproceedings{particle-guidance,
    author = {Corso, Gabriele and Xu, Yilun and De Bortoli, Valentin and Barzilay, Regina  and Jaakkola, Tommi},
    booktitle = {International Conference on Learning Representations},
    pages = {22480--22507},
    title = {Particle Guidance: non-I.I.D. Diverse Sampling with Diffusion Models},
    year = {2024},
}

@InProceedings{fk-steering,
    title = {A General Framework for Inference-time Scaling and Steering of Diffusion Models},
    author = {Singhal, Raghav and Horvitz, Zachary and Teehan, Ryan and Ren, Mengye and Yu, Zhou and Mckeown, Kathleen and Ranganath, Rajesh},
    booktitle = {Proceedings of the 42nd International Conference on Machine Learning},
    pages = {55810--55827},
    year = {2025},
    volume = {267},
    series = {Proceedings of Machine Learning Research},
    month = {Jul},
    day = {13--19},
    publisher = {PMLR},
}

@inproceedings{svdd,
    title = {Derivative-Free Guidance in Continuous and Discrete Diffusion Models with Soft Value-based Decoding},
    author = {Li, Xiner and Zhao, Yulai and Wang, Chenyu and Scalia, Gabriele and Eraslan, Gokcen and Nair, Surag and Biancalani, Tommaso and Ji, Shuiwang and Regev, Aviv and Levine, Sergey and Uehara, Masatoshi},
    booktitle = {Advances in Neural Information Processing Systems},
    pages = {95507--95545},
    publisher = {Curran Associates, Inc.},
    volume = {38, Main Conference},
    year = {2025},
}

@article{discrete-smc,
    title = {Debiasing guidance for discrete diffusion with sequential {Monte Carlo}},
    author = {Lee, Cheuk Kit and Jeha, Paul and Frellsen, Jes and Lio, Pietro and Albergo, Michael Samuel and Vargas, Francisco},
    journal = {arXiv preprint arXiv:2502.06079},
    year = {2025},
}

@inproceedings{discrete-smc-scaling,
    author = {Ou, Zijing and Pani, Chinmay and Li, Yingzhen},
    booktitle = {International Conference on Learning Representations},
    pages = {36740--36775},
    title = {Inference-Time Scaling of Discrete Diffusion Models via Importance Weighting and Optimal Proposal Design},
    year = {2026},
}

@inproceedings{apt,
    author = {Zhang, Leo and Potaptchik, Peter and He, Jiajun and Du, Yuanqi and Doucet, Arnaud and Vargas, Francisco and Dau, Hai-Dang and Syed, Saifuddin},
    booktitle = {International Conference on Learning Representations},
    pages = {90874--90907},
    title = {Accelerated Parallel Tempering via Neural Transports},
    year = {2026},
    note = {Note: This work has previously been published under the title ``Generalised parallel tempering: flexible replica exchange via flows and diffusions''},
}

@inproceedings{source-parallel-tempering,
    title = {Test-Time Guidance for Flow-Based Generative Models via Parallel Tempering on Source Distributions},
    author = {Wang, Shih-Hsin and Keller, Joel A. and Transue, Taos and Brown, Drake Benjamin and Strohmer, Thomas and Wang, Bao},
    booktitle = {Proceedings of the 43rd International Conference on Machine Learning},
    year = {2026},
    series = {Proceedings of Machine Learning Research},
    publisher = {PMLR},
}

@inproceedings{crepe,
    author = {He, Jiajun and Jeha, Paul and Potaptchik, Peter and Zhang, Leo and Hern\'{a}ndez Lobato, Jos\'{e} Miguel and Du, Yuanqi and Syed, Saifuddin and Vargas, Francisco},
    booktitle = {International Conference on Learning Representations},
    title = {CREPE: Controlling diffusion with REPlica Exchange},
    pages = {99464--99494},
    year = {2026},
}

@article{replica-exchange-1,
    title = {Replica Monte Carlo Simulation of Spin-Glasses},
    author = {Swendsen, Robert and Wang, Jian-Sheng},
    year = {1986},
    month = {Nov},
    pages = {2607--2609},
    volume = {57},
    journal = {Physical Review Letters},
}

@inproceedings{replica-exchange-2,
    title = {Markov Chain Monte Carlo Maximum Likelihood},
    author = {Geyer, Charles J},
    booktitle = {Computing science and statistics: Proceedings of the 23rd Symposium on the Interface},
    publisher = {Interface Foundation of North America},
    pages = {156--163},
    year = {1991},
}

@article{replica-exchange-pairs,
    title = {Exchange Monte Carlo method and application to spin glass simulations},
    author = {Hukushima, Koji and Nemoto, Koji},
    journal = {Journal of the Physical Society of Japan},
    volume = {65},
    number = {6},
    pages = {1604--1608},
    year = {1996},
    publisher = {The Physical Society of Japan}
}

@article{parallel-tempering-theory-applications,
    title = {Parallel tempering: Theory, applications, and new perspectives},
    author = {Earl, David J and Deem, Michael W},
    journal = {Physical Chemistry Chemical Physics},
    volume = {7},
    number = {23},
    pages = {3910--3916},
    year = {2005},
    publisher = {The Royal Society of Chemistry}
}

@article{vousden,
    title = {Dynamic temperature selection for parallel tempering in Markov chain Monte Carlo simulations},
    author = {Vousden, Will D and Farr, Will M and Mandel, Ilya},
    journal = {Monthly Notices of the Royal Astronomical Society},
    volume = {455},
    number = {2},
    pages = {1919--1937},
    year = {2016},
    publisher = {Oxford University Press},
}

@inproceedings{rb-graphs,
    author = {Xu, Ke and Boussemart, Frédéric and Hemery, Fred and Lecoutre, Christophe},
    title = {A simple model to generate hard satisfiable instances},
    year = {2005},
    publisher = {Morgan Kaufmann Publishers Inc.},
    booktitle = {Proceedings of the 19th International Joint Conference on Artificial Intelligence},
    pages = {337--342},
}

@article{ba-graphs,
    title = {Emergence of Scaling in Random Networks},
    author = {Albert-László Barabási and Réka Albert},
    journal = {Science},
    volume = {286},
    number = {5439},
    pages = {509--512},
    year = {1999},
}

@misc{gurobi,
    author = {{Gurobi Optimization, LLC}},
    title = {Gurobi Optimizer Reference Manual},
    year = {2026},
}

@article{kamis,
    author = {Sebastian Lamm and Peter Sanders and Christian Schulz and Darren Strash and Renato F. Werneck},
    title = {Finding near-optimal independent sets at scale},
    journal = {J. Heuristics},
    volume = {23},
    number = {4},
    pages = {207--229},
    year = {2017},
}
\bibliographystyle{iclr2027_conference}

\newpage
\appendix

\section{Energy Functions}
\label{sec:energy-functions}

Following \citet{diffuco, sdds}, we used these energy functions in our experiments:

\begin{align*}
    \text{\ac{MIS}:}   \quad & H_\mathcal{G}(\bm{x}) = -A \sum_{i=1}^{|V|} \bm{x}_i + B \sum_{(i,j) \in E} \bm{x}_i \cdot \bm{x}_j \\
    \text{\ac{MDS}:}   \quad & H_\mathcal{G}(\bm{x}) = A \sum_{i=1}^{|V|} \bm{x}_i + B \sum_{i=1}^{|V|} (1 - \bm{x}_i) \prod_{j \in \mathcal{N}_\mathcal{G}(i)} (1 - \bm{x}_j) \\
    \text{Max cut:}    \quad & H_\mathcal{G}(\bm{x}) = - \sum_{(i,j) \in E} \frac{1 - \sigma_i \sigma_j}{2}, \quad \text{where } \sigma_i = 2 \bm{x}_i - 1 \\
    \text{Max clique:} \quad & H_\mathcal{G}(\bm{x}) =  -A \sum_{i=1}^{|V|} \bm{x}_i + B \sum_{(i, j) \notin E} \bm{x}_i \cdot \bm{x}_j \\
\end{align*}

Here, subscripts refer to indices of the vector, not diffusion time steps.
$\mathcal{N}_\mathcal{G}(i)$ refers to the neighbors of node $i$ in graph $\mathcal{G} = (V, E)$.
$A, B \in \mathbb{R}$ are hyperparameters; all experiments use $A = 1$ and $B = 1.1$.

\section{Hyperparameters}
\label{sec:hyperparameters}

A common choice for parallel tempering is to space temperatures geometrically between a minimum and a maximum temperature, meaning that $\tau_{i+1} / \tau_i = \mathrm{const}$ for all $i \in \{ 1, \dots, R - 1 \}$ \citep{parallel-tempering-theory-applications}.
The minimum temperature is usually set to $\tau_1 = 1$.
We follow these choices here.

For diffusion-related hyperparameters, we chose the same values used by \citet{sdds}.
For comparability to DiffUCO and SDDS, our experiments use the trained models available here: \url{https://github.com/ml-jku/DIffUCO/tree/main/Checkpoints}.
For SDDS, models trained using rKL with RL are used, as they perform better on most \ac{CO} problems and datasets.

\Cref{tab:hyperparams} lists the values chosen for each hyperparameter.

\begin{table}[H]
    \caption{Hyperparameters used in our experiments.}
    \label{tab:hyperparams}
    \centering
    \begin{tabular}{lcc}
        \toprule
        Hyperparameter                & Small Datasets & Large Datasets \\
        \midrule
        Minimum temperature $\tau_1$  & \multicolumn{2}{c}{1.0}         \\
        Maximum temperature $\tau_R$  & 5.0            & 2.5            \\
        Number of temperatures $R$    & \multicolumn{2}{c}{10}          \\
        Number of replicas $N$        & \multicolumn{2}{c}{100}         \\
        Number of diffusion steps $T$ & \multicolumn{2}{c}{18}          \\
        \bottomrule
    \end{tabular}
\end{table}

\section{Hardware}
\label{sec:hardware}

Experiments were performed using an NVIDIA H100 GPU with 80GB of vRAM and an AMD EPYC 9654 96-Core processor.

\end{document}